\documentclass{SCIS2025}

\usepackage{graphicx}
\usepackage{enumitem}
\usepackage[edges]{forest} 
\usepackage{xcolor} 
\usepackage{tikz}
\usetikzlibrary{trees}
\usepackage{hyperref}
\usepackage{xcolor}
\usepackage{colortbl}
\usepackage{booktabs}
\usepackage{array}
\usepackage{amsmath} 

\definecolor{MyFillColor}{rgb}{0.8,0.9,1}   
    
\begin{document}
\ArticleType{REVIEW PAPER}
\Year{2025}
\Month{}
\Vol{}
\No{}
\DOI{}
\ArtNo{}
\ReceiveDate{}
\ReviseDate{}
\AcceptDate{}
\OnlineDate{}
\AuthorMark{}
\AuthorCitation{}

\title{Shortcut Learning in In-Context Learning: A Survey}

\author[1,2]{Rui Song}{}
\author[1,2]{Yingji Li}{}
\author[3]{Lida Shi}{}
\author[1,4]{Fausto Giunchiglia}{}
\author[1]{Hao Xu}{xuhao@jlu.edu.cn}


\address[1]{College of Computer Science and Technology, Jilin University, Changchun, China}
\address[2]{Key Laboratory of Symbol Computation and Knowledge Engineering of Ministry of Education, Jilin University, China}
\address[3]{College of Computer Science and Technology, Jilin University, Changchun, China}
\address[4]{Department of Information Engineering and Computer Science, University of Trento, Italy}

\abstract{Shortcut learning refers to the phenomenon where models employ simple, non-robust decision rules in practical tasks, which hinders their generalization and robustness. With the rapid development of Large Language Models (LLMs) in recent years, an increasing number of studies have shown the impact of shortcut learning on LLMs. This paper provides a novel perspective to review relevant research on shortcut learning in In-Context Learning (ICL). It conducts a detailed exploration of the types of shortcuts in ICL tasks, their causes, available benchmarks, and strategies for mitigating shortcuts. Based on the corresponding observations, we summarize the unresolved issues in existing research and attempts to outline the future research landscape of shortcut learning. }


\maketitle

\section{Introduction}

In recent years, Large Language Models (LLMs) have emerged as a hotly pursued research direction, with the advent of major language models such as T5~\cite{T5}, LLaMA~\cite{LLaMA}, PaLM~\cite{PaLM}, GPT-3~\cite{GPT3}, Qwen2~\cite{Qwen2}, and GLM~\cite{GLM}. LLMs demonstrate the ability of In-Context Learning (ICL), meaning they can learn from several demonstration examples within a given context without fine-tuning~\cite{dong2022survey}. ICL has been widely confirmed as an emerging capability of LLMs and has led to new usage paradigms for LLMs~\cite{WeiTBRZBYBZMCHVLDF22, SchaefferMK23}. However, an increasing number of studies have shown that LLMs are susceptible to shortcut learning in ICL and perform poorly in many aspects~\cite{DuHZTH24}.

\begin{figure}
    \centering
    \includegraphics[width=0.45\linewidth]{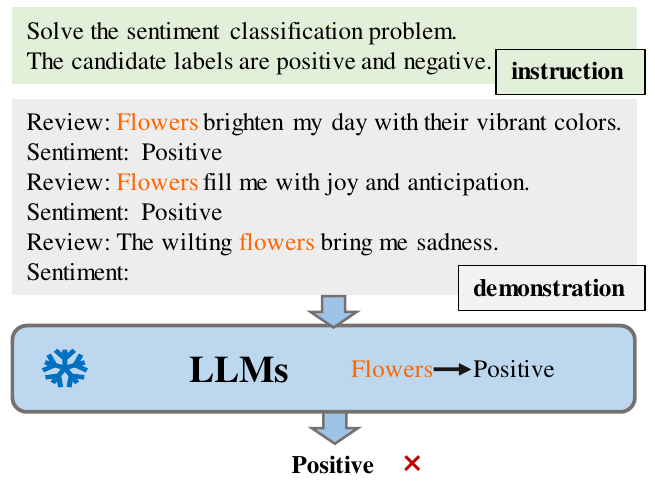}
    \caption{An example of shortcut learning in ICL.}
    \label{fig:exmaple}
\end{figure}

Shortcuts refer to decision rules that perform well on standard benchmarks but fail to transfer to more challenging testing conditions. Shortcut learning is the process by which a model exploits these decision rules during execution, leading to non-robust output results~\cite{GeirhosJMZBBW20}. In the literature, shortcut learning is also referred to as superficial/spurious correlations~\cite{ZhouX0A0H24}, the Clever Hans effect~\cite{lapuschkin2019unmasking}, and various types of biases\footnote{In our survey, bias refers to systematic errors within LLMs, rather than cultural or social biases that are studied in the field of LLM safety and fairness~\cite{Zheng0M0H24}.}, such as learning bias~\cite{DuHZTH24}, label bias~\cite{FeiHCB23}, selection biases~\cite{WeiWHC24}, co-occurrence bias~\cite{KangC23}, and so forth. In practice, if LLMs establish shortcut `Flowers$\rightarrow$Position' based on given samples in Figure~\ref{fig:exmaple}, predictions may fail when confronted with `Flowers$\rightarrow$Negative' samples. Apart from this mapping from lexicon to specific labels, shortcuts also exist widely in various forms across different natural language processing tasks. For example, in Natural Language Inference (NLI)~\cite{ZhouB20}, Question-Answering (QA)~\cite{SenS20}, and Machine Reading Comprehension (MRC)~\cite{LaiZFHZ21}, LLMs have a preference for overlapping parts of two input branches. These unhealthy shortcuts have been widely proven to be harmful to the robustness~\cite{WangSY022}, generalization~\cite{0003L21}, safety~\cite{Safety2024}, morality~\cite{LiuMTJ24}, as well as fairness~\cite{Yingji23}, and may lead to hallucinations in LLMs~\cite{JiLFYSXIBMF23}. 

Given the rapid development of LLMs and the impact of shortcut learning on them, our survey aims to systematically summarize, discuss, and consolidate the existing research, and to sensitize the community to the current advancements. \textbf{Different from previous reviews on shortcut learning}~\cite{Xanh22, DuHZTH24, DograVKWSI24, Geetanjali2024}, our survey focuses on LLMs, while prior research primarily centered on Lightweight Pre-trained Language Models (LPLMs) such as BERT~\cite{DevlinCLT19} and RoBERTa~\cite{RoBERTa}. Considering the differences in the training and application of LLMs, we deem it necessary to reassess the types of shortcuts, the causes of shortcut learning, benchmarks, and mitigation strategies, and to highlight the unresolved issues surrounding shortcut learning in LLMs. Additionally, we have summarized different types of shortcuts and their related research tasks, which can assist researchers in clarifying research directions and quickly identifying application scenarios. The organization of sections is as follows.

\begin{figure}
    \begin{tikzpicture}[
        every node/.style={draw, rounded corners, minimum height=1cm, minimum width=2.5cm, align=center, font=\small},
        link/.style={-, thick}
    ]
    \node (root)[minimum height=0.5cm, minimum width=1.5cm, rotate=90] at ($(0, 1)$) {Shortcut Learning in ICL};
    \node (type)[minimum height=0.5cm, minimum width=1.5cm, rotate=90] at ($(1.2, 8.2)$) {Types §~\ref{sec:type}};
    \node (cause)[minimum height=0.5cm, minimum width=1.5cm, rotate=90] at ($(1.2, 5.5)$) {Causes §~\ref{sec:cause}};
    \node (benchmarks)[minimum height=0.5cm, minimum width=1.5cm, rotate=90] at ($(1.2, 1)$) {Benchmarks §~\ref{sec:benchmark}};
    \node (matigation)[minimum height=0.5cm, minimum width=1.5cm, rotate=90] at ($(1.2, -3)$) {Matigation §~\ref{sec:mitigation}};
    \node (rethinking)[minimum height=0.5cm, minimum width=1.5cm, rotate=90] at ($(1.2, -6.3)$) {Rethinking §~\ref{sec:thinking}};
    \node (instinctive)[minimum height=0.8cm, minimum width=3cm] at ($(3.6, 8.8)$) {Instinctive Shortcuts \\ §~\ref{sec:Instinctive}};
    \node (acquired)[minimum height=0.8cm, minimum width=3cm] at ($(3.6, 7.7)$) {Acquired Shortcuts \\ §~\ref{sec:Acquired}};
    \node (LLMstraining)[minimum height=0.8cm, minimum width=3cm] at ($(3.6, 6.6)$) {LLMs Training \\ §~\ref{sec:LLMstraining}};
    \node (demonstrations)[minimum height=0.5cm, minimum width=3cm] at ($(3.9, 5.5)$) {Skewed Demonstrations \\ §~\ref{sec:demonstrations}};
    \node (size)[minimum height=0.8cm, minimum width=3cm] at ($(3.6, 4.4)$) {LLMs Size \\ §~\ref{sec:size}};
    \node (dataset)[minimum height=0.8cm, minimum width=3cm] at ($(3.6, 2.7)$) {Datasets \\ §~\ref{sec:Datasets}};
    \node (task)[minimum height=0.8cm, minimum width=3cm] at ($(3.6, 1)$) {Tasks \\ §~\ref{sec:tasks}};
    \node (metrics)[minimum height=0.8cm, minimum width=3cm] at ($(3.6, -0.5)$) {Metrics \\ §~\ref{sec:Metrics}};
    \node (Data-centric)[minimum height=0.8cm, minimum width=3cm] at ($(3.6, -1.8)$) {Data-centric \\ §~\ref{sec:Data-centric}};
    \node (Model-centric)[minimum height=0.8cm, minimum width=3cm] at ($(3.6, -3)$) {Model-centric \\ §~\ref{sec:Model-centric}};
    \node (Prompt-centric)[minimum height=0.8cm, minimum width=3cm] at ($(3.6, -4.2)$) {Prompt-centric \\ §~\ref{sec:Prompt-centric}};

    \node (different)[minimum height=0.8cm, minimum width=3cm] at ($(3.6, -5.5)$) {Differences from  \\ LPLMs §~\ref{sec:Prompt-centric}};
    \node (future)[minimum height=0.8cm, minimum width=3cm] at ($(3.6, -7)$) {Future Directions \\ §~\ref{sec:Prompt-centric}};
    
    \node (4Instinctive)[minimum height=0.8cm, minimum width=7cm, fill=MyFillColor] at ($(10.7, 8.8)$)     {\parbox{10cm}{\textbf{Vanilla-label Bias}~\cite{ZhaoWFK021, HoltzmanWSCZ21, Thomas2023, KangC23, PanG0C23, WangMWZC23, Zheng0M0H24, KossenGR24}, \textbf{Context-label Bias}~\cite{ZhaoWFK021, LuBM0S22, Sclar0TS24, WangLCCZLCKLLS24, PezeshkpourH24, WeiWHC24}, \textbf{Domain-label Bias}~\cite{RazeghiL0022, Thomas2023, FeiHCB23, SiFJFC023, balepur2024artifacts, ZhangLWWCJLR24}, \textbf{Reasoning-label Bias}~\cite{Xuansheng2024, Yuan2024, Liu0LDGLZ24, LiZWFRC24, JuCY0DZL24, Congzhi2024, yamin2024failure, WangCWSL024, Wan0YY0HJL24}.}};
    
    \node (4Acquired)[minimum height=0.8cm, minimum width=9cm, fill=MyFillColor] at ($(10.7, 7.7)$)     {\parbox{10cm}{\textbf{Lexicon}~\cite{TangKH023, si2023spurious, SunXLJCZ24, yuan2024llms, Yuqing2024, pacchiardi2024leaving}, \textbf{Concept}~\cite{ZhouX0A0H24, Yuqing2024}, \textbf{Overlap}~\cite{LevyRG23, si2023spurious, SunXLJCZ24, yuan2024llms}, \textbf{Position}~\cite{LevyRG23, yuan2024llms, LiuLHPBPL24}, \textbf{Text Style}~\cite{TangKH023, yuan2024llms, Yuqing2024}, \textbf{Group Dynamics}~\cite{ZhaoWFK021, Karan2023, balepur2024artifacts, Pengrui2024}.}};

    \node (4LLMstraining)[minimum height=0.8cm, minimum width=9cm, fill=MyFillColor] at ($(10.7, 6.6)$)     {\parbox{10cm}{\textbf{Pretraining}~\cite{han2022prototypical, RazeghiL0022, Thomas2023, KangC23, QiLHWLWL23, FeiHCB23, JiangZLZL23, Yuqing2024, JuCY0DZL24, Congzhi2024}, \textbf{Instruction Tuning}~\cite{SunXLJCZ24}. }};
    \node (4demonstrations)[minimum height=0.8cm, minimum width=9cm, fill=MyFillColor] at ($(11, 5.5)$)     {\parbox{9.5cm}{LLMs captures spurious associations of shortcuts with labels in demonstrations~\cite{TangKH023, si2023spurious, LevyRG23}.}};
    \node (4size)[minimum height=0.8cm, minimum width=9cm, fill=MyFillColor] at ($(10.7, 4.4)$)     {\parbox{10cm}{Larger-scale LLMs are more likely to exploit shortcuts~\cite{RazeghiL0022, KangC23, PanG0C23, Yuqing2024, Sclar0TS24, TangKH023, balepur2024artifacts, yuan2024llms, Pengrui2024}. }};

    \node (4benchmarks)[minimum height=0.8cm, minimum width=9cm, fill=MyFillColor] at ($(10.7, 3.3)$)     {\parbox{10cm}{ Using or modifying the existing dataset~\cite{ZhaoWFK021, FeiHCB23, Karan2023, TangKH023, si2023spurious, SunXLJCZ24, ZhouX0A0H24}. }};

    \node (4benchmarks2)[minimum height=0.8cm, minimum width=9cm, fill=MyFillColor] at ($(10.7, 2.1)$)     {\parbox{10cm}{\textbf{ConvRe}~\cite{QiLHWLWL23}, \textbf{ShortcutQA}~\cite{LevyRG23}, \textbf{Shortcut Maze}~\cite{Yuqing2024}, \textbf{Shortcut Suite}~\cite{yuan2024llms}, \textbf{EUREQA}~\cite{LiZWFRC24}, \textbf{MMLU-Pro+}~\cite{Saeid2024}, \textbf{ReWild}~\cite{Yuan2024}.  }};

    \node (4tasks)[minimum height=0.8cm, minimum width=9cm, fill=MyFillColor] at ($(10.7, 1)$)     {\parbox{10cm}{Text Classificaion, Information Extraction, Natural Language Inference, Cloze, Question-Answering, Reasoning.}};

    \node (4metrics)[minimum height=0.8cm, minimum width=9cm, fill=MyFillColor] at ($(10.7, -0.5)$)     {\parbox{10cm}{AUC score~\cite{ChenZYM023}, Accuracy and F1 score~\cite{TangKH023, FeiHCB23}, Hits@n~\cite{KangC23, PezeshkpourH24}, Exact Match score~\cite{Congzhi2024, WangMWZC23}, Perfomance Changes~\cite{ZhouX0A0H24, Pengrui2024, Yuqing2024}, Fluctuation Rate~\cite{WeiWHC24}, Conflict Rate~\cite{Zheng0M0H24}, Shortcut Selection Ratio~\cite{Saeid2024}.}};

    \node (4Data-centric)[minimum height=0.5cm, minimum width=9cm, fill=MyFillColor] at ($(10.7, -1.8)$)     {\parbox{10cm}{\textbf{Resampling} and \textbf{retraining}~\cite{KangC23, ZhouX0A0H24, Ryumei2024}.}};
    \node (4Model-centric)[minimum height=0.8cm, minimum width=9cm, fill=MyFillColor] at ($(10.7, -3)$)     {\parbox{10cm}{\textbf{Model pruning}~\cite{JuCY0DZL24, YangKCLJ24, ali2024mitigating, Hanzhang2024}, \textbf{Calibration}~\cite{ZhaoWFK021, HoltzmanWSCZ21, han2022prototypical, FeiHCB23, JiangZLZL23, ZhouWPMCHR24, JangJKJY24, Zheng0M0H24, Yufeng24}. }};
    \node (4Prompt-centric)[minimum height=0.8cm, minimum width=9cm, fill=MyFillColor] at ($(10.7, -4.2)$)     {\parbox{10cm}{ \textbf{Shortcut-based method}~\cite{WangMWZC23, ZhouX0A0H24}, \textbf{Instruction format-based method}~\cite{SiFJFC023, PezeshkpourH24, WangLCCZLCKLLS24, SunXLJCZ24, QiLHWLWL23, yuan2024llms, Congzhi2024, yamin2024failure}, \textbf{Prompt search-based method}~\cite{LuBM0S22, ChenZYM023, Gonen0BSZ23}.}};
    \node (4different)[minimum height=0.8cm, minimum width=9cm, fill=MyFillColor] at ($(10.7, -5.5)$)     {\parbox{10cm}{ Shortcut Types~§\ref{sec:diff1}, Mitigation Strategies~§\ref{sec:diff2}, Shortcut Search~§\ref{sec:diff3}. }};
    \node (4future)[minimum height=0.8cm, minimum width=9cm, fill=MyFillColor] at ($(10.7, -7)$)     {\parbox{10cm}{ More Robust Evaluation Benchmark~§\ref{sec:future1}, More Shortcut Related Tasks~§\ref{sec:future2}, More Interpretability~§\ref{sec:future3}, More Discussion in Shortcut Unknown Scenarios~§\ref{sec:future4}, Decoupling of Instinctive and Acquired Shortcuts~§\ref{sec:future5}, Exploration of Multiple Shortcut Coexistence~§\ref{sec:future6}. }};
    
    \draw[link] (root.south) -- ++(0.3,0) |- (type.north);
    \draw[link] (root.south) -- ++(0.3,0) |- (cause.north);
    \draw[link] (root.south) -- ++(0.3,0) |- (benchmarks.north);
    \draw[link] (root.south) -- ++(0.3,0) |- (matigation.north);
    \draw[link] (root.south) -- ++(0.3,0) |- (rethinking.north);
    
    \draw[link] (type.south) -- ++(0.3,0) |- (instinctive.west);
    \draw[link] (type.south) -- ++(0.3,0) |- (acquired.west);
    \draw[link] (cause.south) -- ++(0.3,0) |- (LLMstraining.west);
    \draw[link] (cause.south) -- ++(0.3,0) |- (demonstrations.west);
    \draw[link] (cause.south) -- ++(0.3,0) |- (size.west);
    \draw[link] (benchmarks.south) -- ++(0.3,0) |- (dataset.west);
    \draw[link] (benchmarks.south) -- ++(0.3,0) |- (metrics.west);
    \draw[link] (benchmarks.south) -- ++(0.3,0) |- (task.west);
    \draw[link] (matigation.south) -- ++(0.3,0) |- (Data-centric.west);
    \draw[link] (matigation.south) -- ++(0.3,0) |- (Model-centric.west);
    \draw[link] (matigation.south) -- ++(0.3,0) |- (Prompt-centric.west);
    \draw[link] (rethinking.south) -- ++(0.25,0) |- (different.west);
    \draw[link] (rethinking.south) -- ++(0.25,0) |- (future.west);
    
    \draw[link] (instinctive.east) -- ++(0.3,0) |- (4Instinctive.west);
    \draw[link] (acquired.east) -- ++(0.3,0) |- (4Acquired.west);
    \draw[link] (LLMstraining.east) -- ++(0.3,0) |- (4LLMstraining.west);
    \draw[link] (demonstrations.east) -- ++(0.2,0) |- (4demonstrations.west);
    \draw[link] (size.east) -- ++(0.3,0) |- (4size.west);
    \draw[link] (dataset.east) -- ++(0.3,0) |- (4benchmarks.west);
    \draw[link] (dataset.east) -- ++(0.3,0) |- (4benchmarks2.west);
    \draw[link] (task.east) -- ++(0.3,0) |- (4tasks.west);
    \draw[link] (metrics.east) -- ++(0.3,0) |- (4metrics.west);
    \draw[link] (Data-centric.east) -- ++(0.3,0) |- (4Data-centric.west);
    \draw[link] (Model-centric.east) -- ++(0.3,0) |- (4Model-centric.west);
    \draw[link] (Prompt-centric.east) -- ++(0.25,0) |- (4Prompt-centric.west);
    \draw[link] (different.east) -- ++(0.25,0) |- (4different.west);
    \draw[link] (future.east) -- ++(0.25,0) |- (4future.west);

    \end{tikzpicture}
    \caption{The organization of this survey.} 
    \label{fig:tree} 
\end{figure}
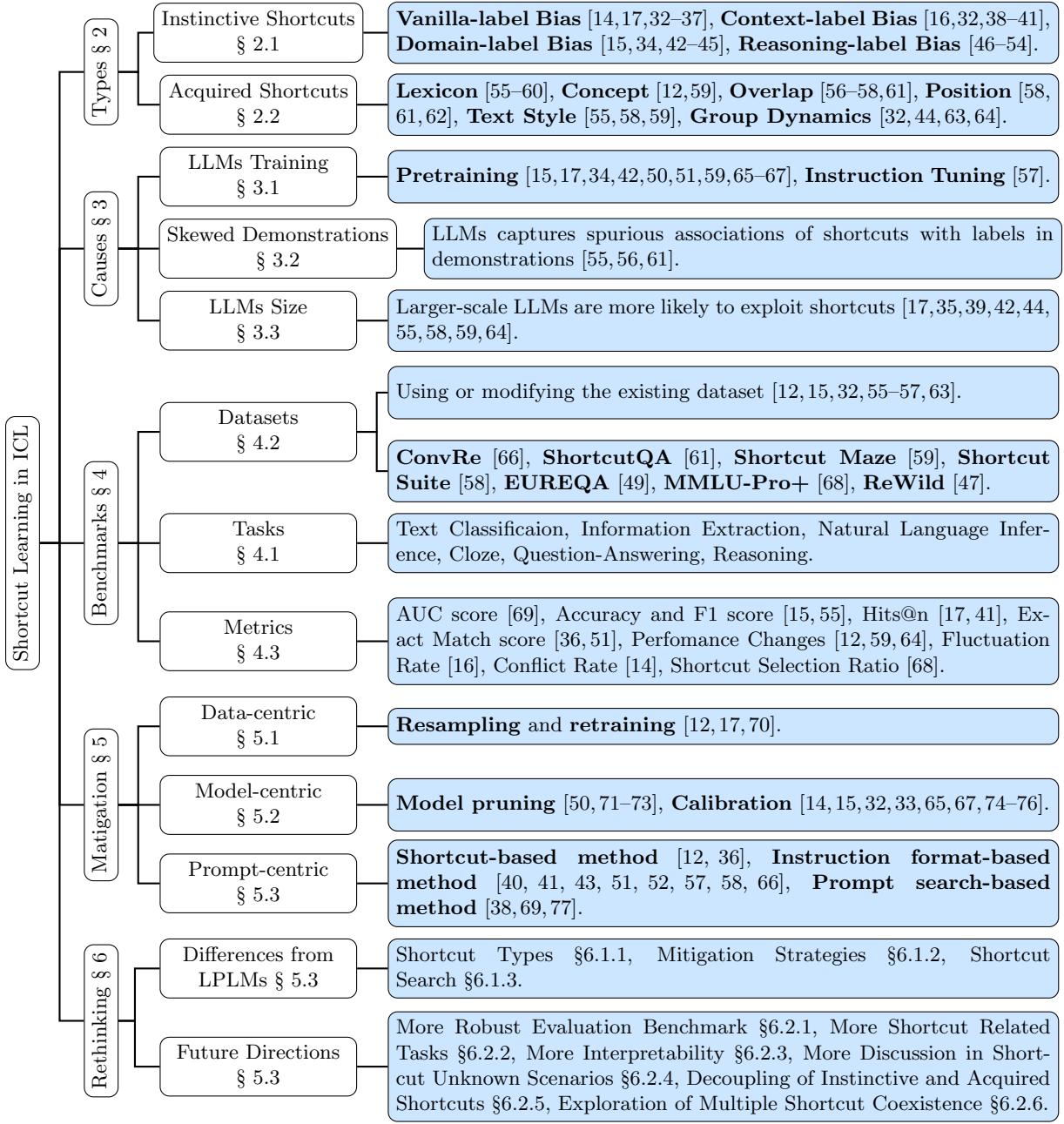

In Section~\ref{sec:type}, we summarize and generalize all the types of shortcuts that have been studied in LLMs. In Section~\ref{sec:cause}, we summarize the possible causes of shortcuts in LLMs. In Section~\ref{sec:benchmark}, we present the benchmarks commonly used in the evaluation tasks for shortcuts in LLMs. In Section~\ref{sec:mitigation}, we provide mitigation strategies for different types of shortcuts. In Section~\ref{sec:thinking}, we conclude and discuss the existing research, and attempt to showcase potential directions for shortcut learning in LLMs. The detailed taxonomy is shown in Figure~\ref{fig:tree}.

\section{Types of Shortcuts}  \label{sec:type}
Compared to~\cite{DuHZTH24}, we extend and summarize various types of shortcuts based on the characteristics of NLP tasks in LLMs. Specifically, we classify shortcuts into \textbf{instinctive shortcuts} and \textbf{acquired shortcuts}, based on whether LLMs rely on prompts to establish superficial correlations with labels or answers. But in reality, we believe that instinctive shortcuts can influence acquired shortcuts to some degree. In other words, instinctive shortcuts may facilitate LLMs in acquiring shortcuts from demonstrations~\cite{PanG0C23}.

\subsection{Instinctive Shortcuts}   \label{sec:Instinctive}
Instinctive shortcuts refer to the inherent preferences of LLMs for texts or patterns outside the examples provided in a given context. These preferences are not influenced by the contextual demonstrations but hinder the effective learning of the input-label relationships presented in those demonstrations. To avoid naming ambiguity, we have adopted the label bias definition proposed by~\cite{FeiHCB23}, which has been widely discussed.

\begin{figure}[h]
    \centering
    \includegraphics[width=0.85\linewidth]{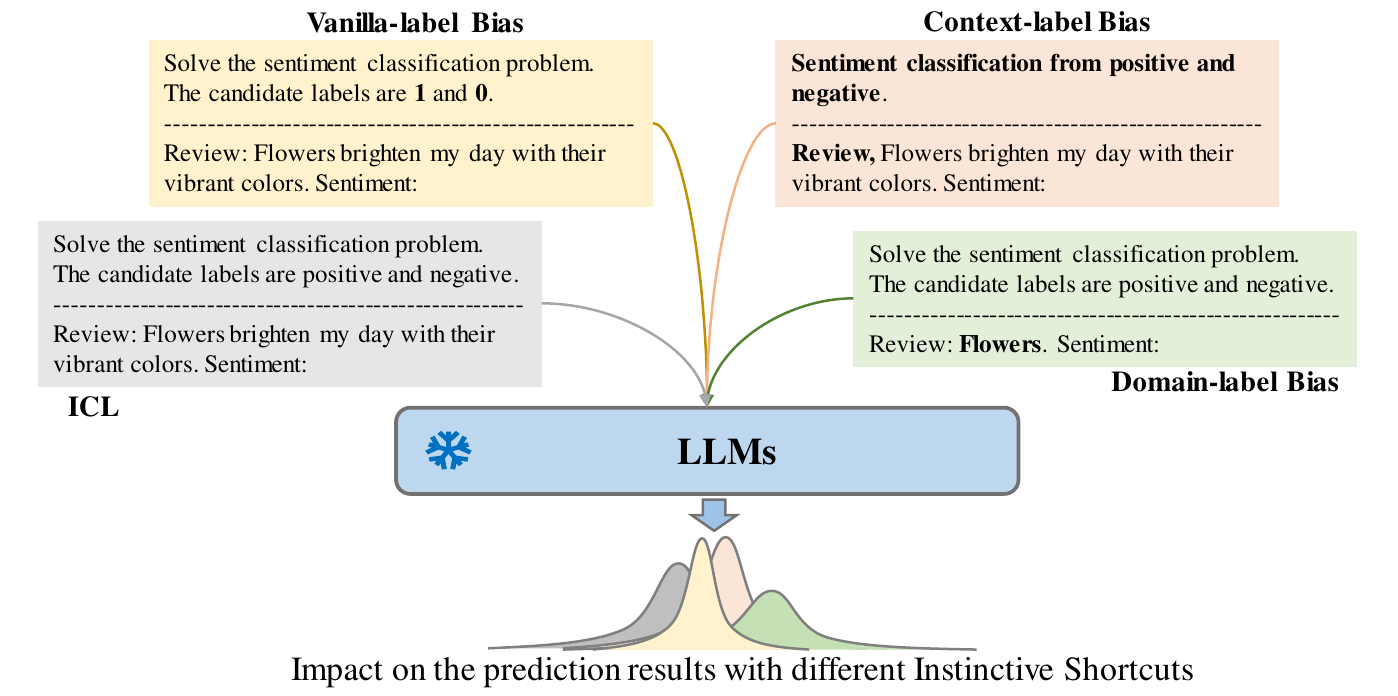}
    \caption{Examples of different instinctive shortcuts. }
    \label{fig:IS-1}
\end{figure}

\begin{itemize}[leftmargin=*,align=left]
    \item \textbf{Vanilla-label Bias} describes the model's uncontextual preference for predicting certain label names. Some studies point out that regardless of the prompt, LLMs tend to assign higher probabilities to specific labels~\cite{ZhaoWFK021, Zheng0M0H24}. Ari et al. demonstrate the existence of Surface Form Competition among different words, leading to a significant decline in prediction performance when low-frequency words are the correct answers~\cite{HoltzmanWSCZ21}. Similarly, furthur studies also confirm the preference of LLMs for high-frequency labels~\cite{Thomas2023, KangC23}. Jane et al. show that replacing labels with meaningless words, such as `@\#\$', leads to a decline in performance compared to using meaningful labels like `ABC'~\cite{PanG0C23} . This preference can sometimes manifest itself in terms of entities, such as LLMs can stubbornly believe that 'Bill Gates' is the 'founder' of 'Microsoft' and ignore the context~\cite{WangMWZC23}. This prior preference is harmful in some cases, resulting in the difficulty for LLMs to completely overcome the predicted preference obtained from the pre-training data and consider all contextual information unequally~\cite{KossenGR24}.

    \item \textbf{Context-label Bias} reveals the influence of context prompt, including factors such as order and format, on prediction outcomes. For example, subtle adjustments to the prompt format can lead to high instability in model performance, such as changing `Passage:\ text' to `Passage text'~\cite{Sclar0TS24}. Moreover, LLMs are also highly sensitive to the order of inputs, whether it be the order of demonstrations or the order of candidate answer options. For example, LLMs are thought to favor options presented at a particular ranking position, such as first or last~\cite{WangLCCZLCKLLS24, PezeshkpourH24, Zheng0M0H24, WeiWHC24}. Zhao et al. also observe that answers closer to the end of the prompt are more likely to be repeated by the model~\cite{ZhaoWFK021}. Lu et al. demonstrate the sequence of presented samples can lead to a vast difference in performance, ranging from nearly achieving the current state-of-the-art to merely performing at the level of random guessing~\cite{LuBM0S22}.

    \item \textbf{Domain-label Bias} describes the dependence on prior knowledge related to the task. Research has indicate that the in-domain words randomly sampled from the dataset are heavily biased towards the domain labels in ICL~\cite{FeiHCB23}. Accordingly, label words also often imply specific task types~\cite{SiFJFC023}. For example, the word `positive' frequently appears in connection with sentiment classification tasks. If `positive' were replaced with a meaningless symbol like `1', LLMs would find it difficult to discern whether the task is sentiment classification or others. This leads to LLMs relying on their semantic priors in ICL predictions, rather than learning from the input-label relationships presented in demonstrations~\cite{JangJKJY24}. This prior can also be observed in QA tasks, where LLMs are able to make correct predictions even without being given options~\cite{balepur2024artifacts}. Additionally, LLMs are also highly sensitive to term frequencies and task probabilities~\cite{RazeghiL0022}. For instance, when asked to decode shift ciphers with various shift levels, GPT-4 scores 50\% or above for the three most common shift numbers (1, 3, and 13), but scores below 3\% for all other shift levels represented by different numbers~\cite{Thomas2023}.
    
    Apart from the semantic priors between texts and labels, entity knowledge within LLMs can also lead to bias~\cite{WangCZCLLYLH22}. For instance, LLMs may assume that `Bill Gates' is the \textit{founder} of `Microsoft', even when given the context `Bill Gates went to Microsoft Building 99' where the relationship between `Bill Gates' and `Microsoft' is \textit{visitor}. Similarly, in relation extraction tasks where the intended relationship to be extracted is unrelated to the existing relationships in the input, LLMs may still extract these unrelated relationships~\cite{ZhangLWWCJLR24}. 

    \begin{figure}[h]
        \centering
        \includegraphics[width=0.9\linewidth]{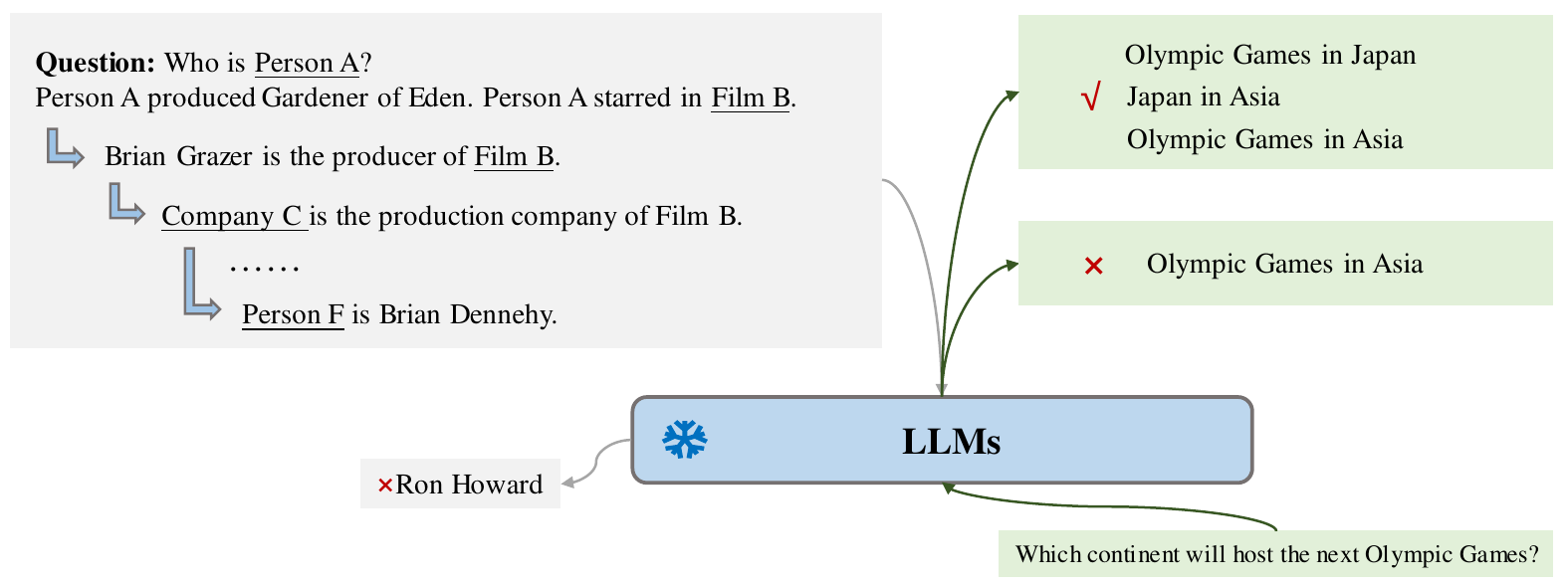}
        \caption{Examples of different reasoning-label bias. }
        \label{fig:IS-2}
    \end{figure}
    
    \item \textbf{Reasoning-label Bias} is a newly summarized shortcut type in this paper compared with~\cite{FeiHCB23}. It indicates that LLMs tend to get superficial answers directly from the provided examples rather than reasoning the underlying patterns~\cite{yu2024reasonagain} and true causal effect~\cite{Congzhi2024}, especially in complex reasoning (eg. compositional reasoning~\cite{LiJ0SL024, WangCWSL024}, logical reasoning~\cite{Wan0YY0HJL24}) and summarization tasks involving long contexts~\cite{Xuansheng2024, Yuan2024}. Reasoning-label bias is affected by several factors, including the absence and error of inference logic~\cite{Liu0LDGLZ24}, the potential association between the head entity and the entity to be predicted~\cite{LiZWFRC24, JuCY0DZL24}, and the order of causal relations in the narrative~\cite{yamin2024failure}. For example, in the reasoning example shown in Figure~\ref{fig:IS-2}, the complex reasoning of multiple hops may cause the reasoning cues of LLMs to be lost. In the case about `Olympics', LLMs may directly build a false inference path that skips `Japan' and still outputs the desired result. At first glance, reasoning-label bias seems to rely on demonstrations, but essentially, it represents a tendency of LLMs to take shortcuts when solving complex problems. Therefore, we categorize it as an instinctive shortcut. 
    
\end{itemize}

\subsection{Acquired Shortcuts} \label{sec:Acquired}
Acquired shortcuts refer to the shortcuts present in demonstrations, which LLMs capture as non-robust patterns through observation of these demonstrations. These patterns are then utilized by LLMs for inference, leading to a decline in performance. 

\begin{figure}[h]
    \centering
    \includegraphics[width=1\linewidth]{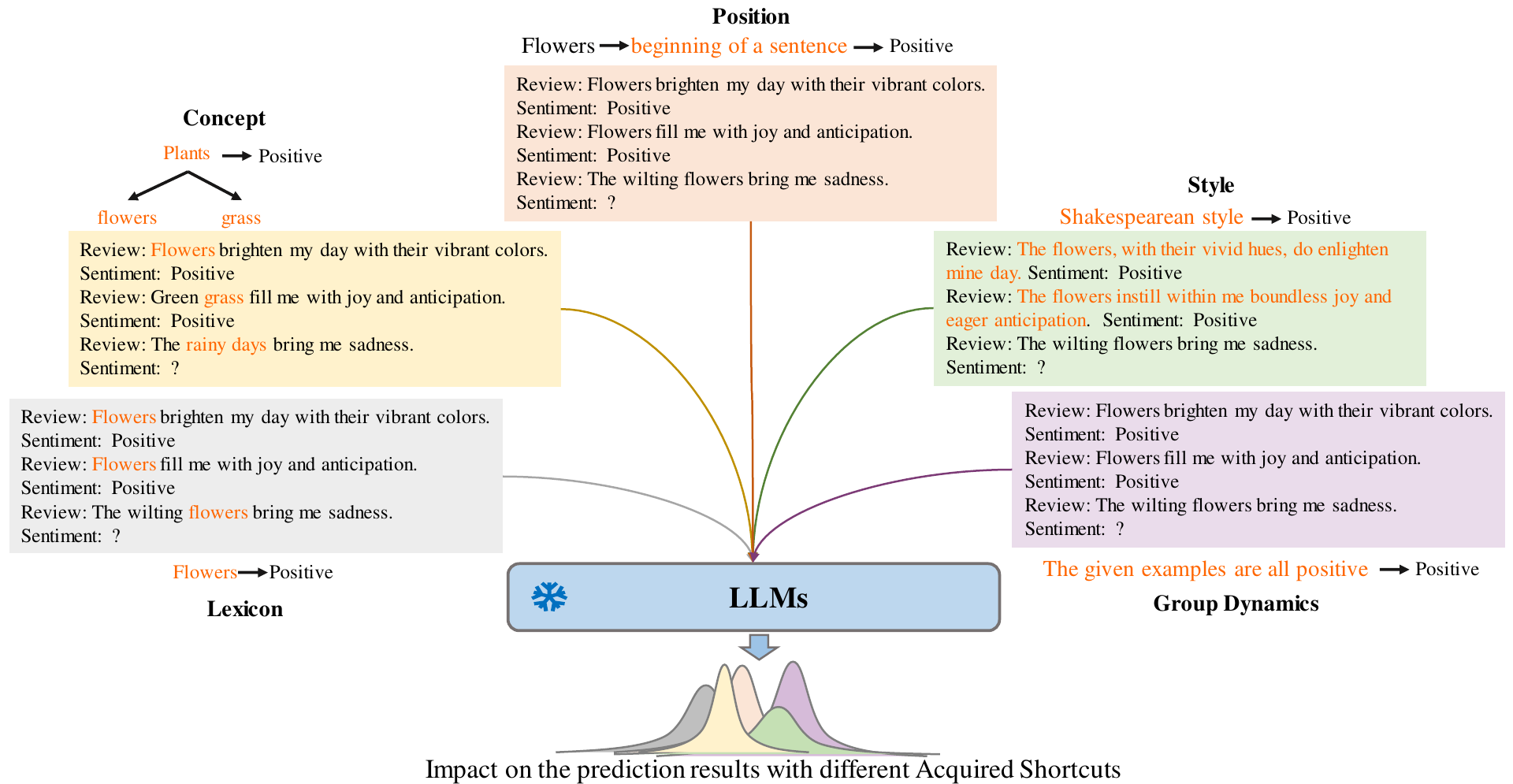}
    \caption{Examples of different acquired shortcuts. }
    \label{fig:AS}
\end{figure}

\begin{itemize}[leftmargin=*,align=left]
    \item \textbf{Lexicon}. When spurious correlation are established between certain lexical features and specific labels, the corresponding lexicon may then become a shortcut. Arbitrary text can serve as lexicon shortcuts, including letters, signs, numbers, negation words, adverbs of degree, or even recurring sentences~\cite{TangKH023}. Among them, negation words have garnered the most attention from researchers, as LLMs often provide contradictory predictions due to these words~\cite{SunXLJCZ24, yuan2024llms, Yuqing2024} But studies have indicated that LLMs are more likely to rely on n-grams and content words rather than others, such as stop words~\cite{si2023spurious, pacchiardi2024leaving}. 

    \item \textbf{Concept}. Concept shortcut refers to the phenomenon where the concepts of certain texts in the demonstrations are disproportionately associated with the labels~\cite{ZhouX0A0H24}. For example, if sentences containing cities are associated with negative emotions and samples containing countries are also associated with positive emotions, when LLMs encounter new city words in the test sample, they may predict it as having negative emotions~\cite{Yuqing2024}. 

    \item \textbf{Overlap}. Overlap shortcuts often emerge in tasks that involve two branches, such as NLI and QA. In NLI, overlap between the premise and the hypothesis can be a strong indicator of \textit{Entailment}~\cite{si2023spurious, SunXLJCZ24, yuan2024llms}. In extractive QA, words that appear in both the question and the background may be used as anchors. LLMs tend to extract answers around these anchors to reduce the error rate~\cite{LevyRG23}. 

    \begin{figure}[h]
        \centering
        \includegraphics[width=1\linewidth]{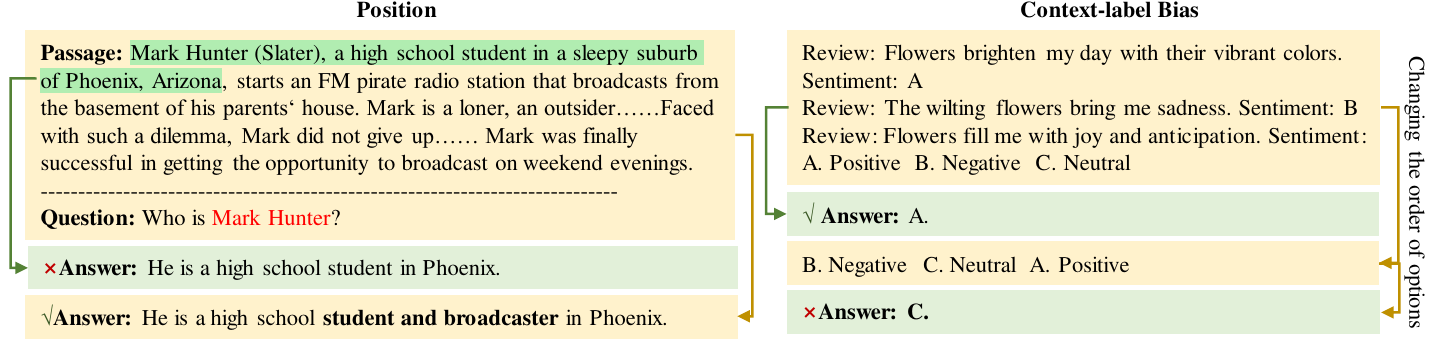}
        \caption{Position Shortcut versus Context-label Bias. }
        \label{fig:position}
    \end{figure}

    \item \textbf{Position}. LLMs use positional information as a basis for prediction. For example, if the answer is located at the beginning or end of the text, the model is less likely to make an error~\cite{LevyRG23, LiuLHPBPL24}. Conversely, Yu et al. show that adding content unrelated to the answer at the beginning of a sentence can degrade the prediction performance of LLMs~\cite{yuan2024llms}. But sometimes, LLMs can briefly read the opening segment to predict the answer without needing to read the entire document~\cite{Tianle24}. Unlike the positional preference observed in context-label bias, \textbf{Position} refers to the model's bias resulting from the location of a specific segment of text within the demonstration. The former is caused by the sensitivity of LLMs to the formats of prompts irrespective of the text's specific content, whereas the latter involves the concrete content of the examples as shown in Figure~\ref{fig:position}.

    \item \textbf{Text Style}. The writing style of the text is also a possible shortcut~\cite{QiCZLLS21}. Some studies demonstrate that shifts in presentation style can have an impact on performance, such as when standard English is rewritten into Shakespearean English~\cite{TangKH023, yuan2024llms}. LLMs can also capture the correlation between style and labels, leading to a decline in the generalization ability of ICL~\cite{Yuqing2024}. 

    \item \textbf{Group Dynamics}. It refers to the interactions, relationships, and influences among members within a group~\cite{cartwright1968group}. Similarly, in the decision-making process, LLMs may evaluate a choice based on other available options. For instance, if other options are even numbers, the model may tend to lean towards an odd number choice~\cite{balepur2024artifacts}. This type of shortcut can also be associated with the A-Not-B error in the field of cognitive science~\cite{PopickDKR11}. If `A' appears more frequently in the demonstrations, the model is more inclined to predict `A', even if `B' is correct~\cite{ZhaoWFK021, Karan2023, Pengrui2024}.

\end{itemize}

\section{Causes of Shortcut Learning}  \label{sec:cause}
In light of the impact of shortcut learning, some methods attempt to provide explanations for the causes of shortcut learning in ICL. This paper divides the causes of shortcut learning in LLMs into three parts: LLMs training, skewed demonstrations, and LLMs size. 

\subsection{LLMs Training} \label{sec:LLMstraining}

Instinctive shortcuts reveal the inherent biases of LLMs. The causes of inherent biases are diverse and closely related to the training process of LLMs, involving factors such as data distribution and instruction tuning. This makes it difficult for LLMs to fully overcome the prediction biases acquired from pre-training data and to consider all contextual information equally during the ICL process~\cite{KossenGR24}.

\begin{itemize}[leftmargin=*,align=left]
    \item \textbf{Pretraining}. Some studies suggest that instinctive shortcuts originate from the data distribution during the pretraining process~\cite{RazeghiL0022, Thomas2023}, which leads to a bias in the predictions of LLMs towards high-frequency and frequently co-occurred words~\cite{KangC23, JuCY0DZL24}. Furthermore, Yu el al. utilize SHAP analysis~\cite{LundbergL17} to confirm that high-frequency words that have spurious correlations with a certain label contribute more significantly to the model's predictions for that label~\cite{Yuqing2024}. Additionally, if the frequency of a particular LLMs task is low, its prediction results will also be inferior compared to those of high-frequency tasks~\cite{Thomas2023}. Based on the observations mentioned above, further work implicitly assume that there is a shift in the label marginal of the ICL predictive distribution~\cite{han2022prototypical, FeiHCB23}, and Jiang et al. theoretically proves that the emergence of shortcuts stems from label shifting caused by the contextual label marginal typically deviating from the true label distribution~\cite{JiangZLZL23}. 

    In addition to the above effects on the preference for certain labels and words, a biased training corpus can also hamper the inference ability of LLMs. The bias present in the training corpus can cause ICL to rely on irrelevant text ranges in the prompts and generate incoherent chains of thoughts, which impairs the logical reasoning ability of the model~\cite{Congzhi2024}. Rare behavioral patterns in the training data, such as converse relations, can also confuse LLMs, causing them to rely on shortcuts when making judgments~\cite{QiLHWLWL23}.

    \item \textbf{Instruction Tuning}. Another perspective holds that the instinctive shortcuts in LLMs originate from the process of instruction tuning, during which the LLMs learn spurious correlations within the instructions~\cite{SunXLJCZ24}. In complex multitasking scenarios, LLMs can also learn more intricate associations between tasks, features, and labels during the fine-tuning process. As a result, LLMs can exploit these spurious correlations simply by being provided with useful task information~\cite{PanG0C23}. But research on shortcuts for instruction tuning is still relatively few.
\end{itemize}

\subsection{Skewed Demonstrations} \label{sec:demonstrations}
Corresponding to LLMs training, skewed demonstrations are primarily responsible for the generation of acquired shortcuts, as LLMs learn from and reason based on the models presented in the demonstrations. If the demonstrations are skewed, the corresponding reasoning results will also be affected. Many studies have confirmed this by adding specific shortcut triggers to the demonstrations, where the shortcuts in the demonstrations caused a decline in the performance of LLMs~\cite{TangKH023, si2023spurious, LevyRG23}. 

\subsection{LLMs Size} \label{sec:size}
While larger LLMs exhibit exceptional capabilities in semantic understanding and reasoning, numerous studies have shown that increasing the LLMs' scale does not mitigate shortcut learning~\cite{KangC23, Yuqing2024, Sclar0TS24}. Rather, even larger LLMs are more prone to exploiting shortcuts, whether they are instinctive shortcuts or acquired shortcuts~\cite{RazeghiL0022, TangKH023, balepur2024artifacts, yuan2024llms, Pengrui2024}. Although not directly stated, Pan et al. provide some explanation for this phenomenon: larger LLMs can utilize more demonstrations to enhance their ability to learn new input-label mappings, whereas smaller LLMs can not~\cite{PanG0C23}. Consequently, when there are more shortcuts in the context, larger LLMs are more susceptible to being influenced by these shortcuts.

\section{Benchmarks of Shortcut Learning}  \label{sec:benchmark}
ICL can encompass many of NLP tasks, targeting different types of shortcuts. Therefore, we deem it necessary to show existing tasks and evaluation benchmarks to support related research on different situations. Specifically, we summarize the the NLP tasks related to shortcut learning, the datasets used in existing research, along with the corresponding quantitative metrics.

\begin{table*}[htbp]
  \centering
  \setlength{\tabcolsep}{3pt} 
  \small
  \begin{tabular}{m{6em}<{\centering}|m{6em}<{\centering}|m{6em}<{\centering}|m{6em}<{\centering}|m{6em}<{\centering}|m{6em}<{\centering}|m{6em}<{\centering}}
 \toprule
     & \textbf{Text Classification} & \textbf{Information Extraction} & \textbf{Natural Language Inference} & \textbf{Cloze} & \textbf{QA} & \textbf{Reasoning} \\ \hline
    \textbf{Vanilla-label Bias} & \cite{ZhaoWFK021,HoltzmanWSCZ21,PanG0C23,KossenGR24} & \cite{ZhaoWFK021,WangMWZC23} & \cite{PanG0C23,KossenGR24} & \cite{ZhaoWFK021,KangC23} & \cite{HoltzmanWSCZ21, WangMWZC23, Zheng0M0H24} &  {\cite{Thomas2023}} \\ \hline
    \textbf{Context-label Bias} & \cite{ZhaoWFK021,LuBM0S22,Sclar0TS24} & \cite{ZhaoWFK021} & \checkmark     & \cite{ZhaoWFK021} & \cite{WeiWHC24,PezeshkpourH24,Sclar0TS24} & \checkmark \\ \hline
    \textbf{Domain-label Bias} & \cite{SiFJFC023} & \cite{ZhangLWWCJLR24} & \cite{FeiHCB23,SiFJFC023}  & \checkmark & \cite{SiFJFC023,balepur2024artifacts} &  {\cite{RazeghiL0022,Thomas2023}} \\ \hline
    \textbf{Reasoning-label Bias} & $\times$ & $\times$ & $\times$ & $\times$ & $\times$ & \cite{Yuan2024,JuCY0DZL24,Liu0LDGLZ24,yamin2024failure,yu2024reasonagain,WangCWSL024,LiZWFRC24,Congzhi2024,Wan0YY0HJL24} \\ \hline
    \textbf{Lexicon} & \cite{TangKH023,si2023spurious,Yuqing2024,pacchiardi2024leaving} & \cite{TangKH023} & \cite{si2023spurious,yuan2024llms,SunXLJCZ24,pacchiardi2024leaving} &  \checkmark  & \cite{pacchiardi2024leaving} &  {\cite{pacchiardi2024leaving}} \\ \hline
    \textbf{Concept} & \cite{ZhouX0A0H24,Yuqing2024} & \checkmark & \checkmark & \checkmark & \checkmark & \checkmark \\ \hline
    \textbf{Overlap} & $\times$ & $\times$ & \cite{si2023spurious,yuan2024llms,SunXLJCZ24} & $\times$ & \cite{LevyRG23} &  $\bigcirc$ \\  \hline
    \textbf{Position} & \checkmark & \checkmark & \cite{yuan2024llms} & \checkmark &  \cite{LevyRG23, Tianle24} &  \cite{LiuLHPBPL24} \\  \hline
    \textbf{Text Style} & \cite{TangKH023,Yuqing2024} & \checkmark & \cite{yuan2024llms} & \checkmark & \checkmark & \checkmark \\  \hline
    \textbf{Group Dynamics} & \cite{ZhaoWFK021,Karan2023} & \cite{ZhaoWFK021} & \cite{Karan2023} & \checkmark & \cite{Karan2023,Pengrui2024,balepur2024artifacts} & \checkmark \\ \toprule
    \end{tabular}%
  \caption{Different types of shortcuts and related tasks. \checkmark indicates that there may also be shortcut types in the corresponding NLP tasks, but the current research has not discussed them in depth. $\times$ denotes that the shortcut type is not suitable for the task. $\bigcirc$ indicates that shortcut types are suitable for the task in some cases. }
  \label{tab:tasks}%
\end{table*}%

\subsection{Tasks} \label{sec:tasks}
We give five common NLP tasks that explore shortcut learning and related research in Table~\ref{tab:tasks}. Due to the intersection between different NLP tasks, we briefly state the tasks presented in Table~\ref{tab:tasks}.

\begin{itemize}[leftmargin=*,align=left]
    \item \textbf{Text Classification} is the most basic task of NLP~\cite{SongGZTX22}, so it has been widely studied in shortcut learning, including sentiment classification~\cite{Shiyuan2023}, topic classification~\cite{han2022prototypical}, toxicity detection~\cite{FeiHCB23}, and so on. 

    \item \textbf{Information Extraction} aims to extract predefined information from unstructured text, typically tasks including relations, entities, and events~\cite{ZhangLWWCJLR24}. In some cases, relation extraction may degenerate into an relation classification task~\cite{ZengLLZZ14}. But in our taxonomy, information extraction is a multivariate task, such as relation extraction, which needs to obtain both entity pairs and triples of their relations.

    \item \textbf{Natural Language Inference} is a typical task that involves two text branches, which determines whether a statement is relevant to the premise and logically reasonable given the premise. It determines whether the second sentence is an Entailment, Contradiction, or Neutral of the first sentence by judging the relationship between two sentences~\cite{yuan2024llms}. 

    \item \textbf{Cloze} is a context-based prediction task that typically involves removing one or more words from a sentence or paragraph, and then asking the model to predict these removed words based on the context~\cite{ZhangMK23}. 

    \item \textbf{QA}. Research in the field of QA is large and diverse, and QA can be classified into different categories based on factors such as context type, context source, type of reasoning, answer format, and so on~\cite{AbdelNabiAA23}. In terms of task form, reasoning is sometimes presented in the form of QA. But in the QA tasks discussed in this survey, complex reasoning is not required, rather, the focus is on retrieving or selecting relevant information from the text and generating answers. The emphasis of the question answering task is on information retrieval and option selection, not on deriving new conclusions, such as in multiple-choice QA tasks that consider shortcuts like prompt format, option order, etc.~\cite{LuBM0S22}, or in machine reading comprehension tasks that involve overlap shortcuts~\cite{LevyRG23}.

    \item \textbf{Reasoning} is a process of integrating multiple types of knowledge to draw new conclusions about the world~\cite{YuZTW24}, which challenges LLMs' ability to perform mathematical reasoning~\cite{Thomas2023}, causal reasoning~\cite{yamin2024failure}, multi-hop reasoning, and other complex cognitive processes~\cite{LiZWFRC24}. It is important to note that, in some cases, NLI is considered a natural language understanding task~\cite{BowmanAPM15}, while in other cases, it is viewed as a reasoning task~\cite{ClarkTR20}. In our survey, NLI is excluded from reasoning.
    
\end{itemize}
    
Furthermore, we observe the relationship between shortcut types and tasks. Firstly, Reasoning-label Bias is highly coupled with reasoning tasks, which is why it does not manifest in other types of tasks. Similarly, Overlap only appears in tasks that contain different branches, making it mutually exclusive with text classification, information extraction, and cloze tests. In the context of reasoning, if the text contains only one branch, overlap is not applicable, such as in the derivation of arithmetic sequences. Conversely, if the task involves multiple branches, where overlapping content exists between different branches, like in multi-hop question answering, then overlap may have an impact. Additionally, we have also observed the absence of certain shortcut types in specific tasks, such as the impact of Text Style on the cloze ability of LLMs. For certain rare text styles, the cloze ability of LLMs may decline. 

\subsection{Datasets}  \label{sec:Datasets}

Most studies have not developed shortcut evaluation datasets, but instead utilize existing datasets~\cite{ZhaoWFK021, FeiHCB23, Karan2023} or make simple modifications to adapt them to their own shortcut evaluation tasks. The modification relies on researchers' prior knowledge of shortcuts. The most common approach is to add or delete elements from the samples in existing datasets to construct samples that contain shortcuts. For exampple, studies introduce shortcuts by injecting shortcut trigger words, typically a sequence of characters, into the samples, and use this method to evaluate the impact of these shortcuts on ICL~\cite{TangKH023, si2023spurious}. In addition to simple tokens or n-grams, context-free sentences can also be used as triggers~\cite{yuan2024llms}. Alternatively, shuffling or deleting option IDs can be used to evaluate the impact of option order on ICL~\cite{Zheng0M0H24}. Some methods also select a subset of samples that meet certain criteria from existing datasets for experimentation. For example, Sun et al. choose samples containing words like `film' or `movie' for lexicon shortcuts~\cite{SunXLJCZ24}. LLMs can also play a role for dataset modifications, such as extracting concepts from text for concept shortcuts~\cite{ZhouX0A0H24}. 

There are also methods that propose corresponding benchmarks for different types of shortcuts to better evaluate them, as the the datasets mentioned above are not specifically designed for shortcut evaluation. Shortcut Maze is a comprehensive benchmark for text classification that categorizes shortcuts into lexicon, style, and concept~\cite{Yuqing2024}. Similarly, Shortcut Suite includes lexicon, overlap, position, and style shortcuts, but it focuses the research on NLI tasks~\cite{yuan2024llms}. Apart from classification tasks, QA benchmarks are also a common scenario for exploring shortcuts. ShortcutQA performs counterfactual editing of samples by adding or excluding shortcut triggers based on the confidence of the target model to obtain shortcut-related samples. It involves various types of shortcuts, including entity, overlap, position, and more~\cite{LevyRG23}. EUREQA is a complex multi-hop QA dataset designed to reduce semantic relevance by adding deceptive semantic cues that distract attention and are unrelated to the correct answer, and to measure the ability of LLMs to perform extensive chained reasoning processes~\cite{LiZWFRC24}. MMLU-Pro+ introduces various interfering shortcuts into the correct options of reasoning questions across 14 different domains to enhance its ability to evaluate high-order reasoning skills in LLMs~\cite{Saeid2024}. ReWild focuses more on testing LLMs at a fine-grained level in various aspects such as mathematics, logic, and common sense, including the selection and execution of strategies, as well as the tendency to adopt undesirable shortcuts~\cite{Yuan2024}. ConvRe systematically evaluates the ability of LLM to recognize and process inverse relations, determining whether these inverse relations are affected by shortcuts through the mutual matching of knowledge triples with their corresponding textual descriptions~\cite{QiLHWLWL23}. 

\subsection{Metrics} \label{sec:Metrics}

The most commonly used metrics in shortcut evaluation are general evaluation methods for NLP tasks such as accuracy, AUC score~\cite{ChenZYM023} as well as F1 score in classification tasks~\cite{TangKH023, FeiHCB23}, Hits@n in retrieval related tasks~\cite{KangC23, PezeshkpourH24}, and exact match score in MRC tasks~\cite{Congzhi2024, WangMWZC23}. If the degree of shortcut learning in ICL is significant, then the corresponding perfomance will also undergo changes~\cite{ZhouX0A0H24, Pengrui2024, Yuqing2024}. Similarly, Fluctuation Rate is designed to represent the ratio of the change in prediction results to the total after a sample is perturbed~\cite{WeiWHC24}. In other studies, Conflict Rate is also equivalent to Fluctuation Rate~\cite{Zheng0M0H24}. Asgari et al. define the Shortcut Selection Ratio, which is used to quantify the tendency of LLMs to maintain their original choices after introducing distracting items~\cite{Saeid2024}. Overall, the evaluation criteria for shortcuts are closely tied to the specific tasks they belong to, with task-related changes in performance serving as the primary basis for assessment. 

\section{Mitigation of Shortcut Learning}   \label{sec:mitigation}
    
\begin{figure}[h]
    \centering
    \includegraphics[width=1\linewidth]{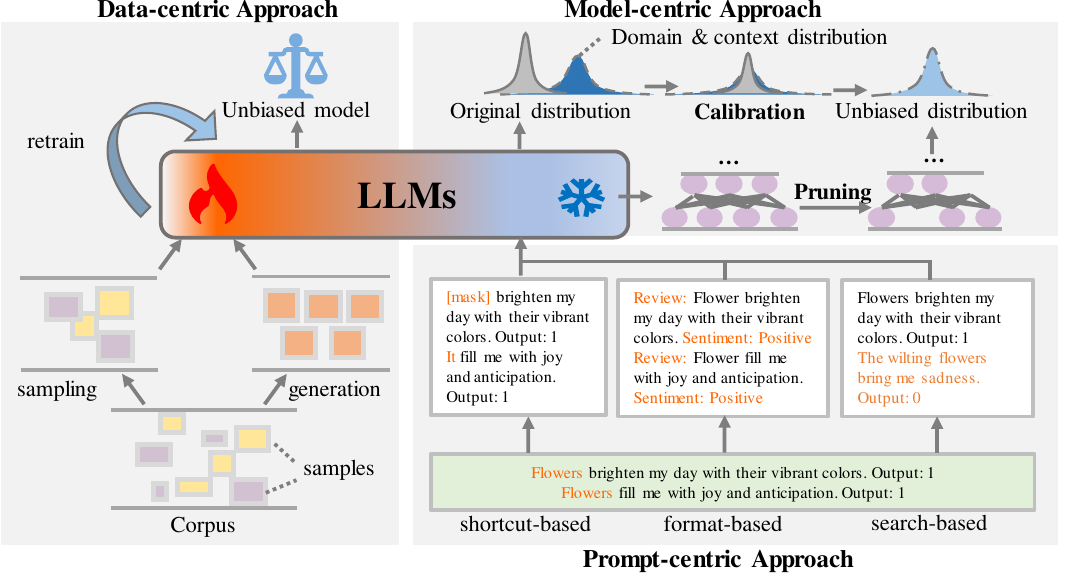}
    \caption{Summary of shortcut mitigation methods. }
    \label{fig:methods}
\end{figure}

Given the widespread existence and dangers of shortcuts, numerous studies have investigated corresponding methods to mitigate them. Based on the differences in the targets addressed by these mitigation methods, we categorize the existing approaches into three types: data-centric, model-centric, and prompt-centric as shown in Figure~\ref{fig:methods}.

\subsection{Data-centric Approach}  \label{sec:Data-centric}
As summarized in Section~\ref{sec:cause}, shortcuts in LLMs are largely related to the distribution of training corpus. Therefore, an intuitive approach is to sample task-relevant examples to obtain balanced training data. Kang et al. alleviate the dependence on shortcuts by filtering out samples with high co-occurrence probabilities and guiding the retraining of LLMs~\cite{KangC23}. Zhou et al. employ counterfactual samples generated by LLMs for upsampling and re-training to mitigate the reliance on shortcuts~\cite{ZhouX0A0H24}. Nakada et al. propose a systematic oversampling method that leverages the capabilities of LLMs to generate high-quality synthetic data for minority groups~\cite{Ryumei2024}. However, the cost of fine-tuning LLMs is too high, and it can also result in catastrophic forgetting~\cite{Chen2024}. Therefore, there are relatively few corresponding studies and focus on LLMs with small parameter numbers.

\subsection{Model-centric Approach} \label{sec:Model-centric}
The model-centric approach mitigates the impact of shortcuts on the results by manipulating neurons and the predictive probability distribution. The main methods can be divided into those based on model pruning and those based on calibration.

\begin{itemize}[leftmargin=*,align=left]
    \item \textbf{Model pruning} alleviates the impact of shortcuts by removing biased neurons. This operation can be supported by research on the relationship between neurons and knowledge storage, where neurons are the basic units for storing knowledge and patterns in LLMs~\cite{zhao2024towards}. In this process, some commonly used post-hoc interpretable methods, such as Integrated Gradients~\cite{SundararajanTY17}, are used to locate the neurons activated during shortcut learning. By removing these neurons, LLMs can be prompted to explore alternative and correct reasoning paths~\cite{JuCY0DZL24, YangKCLJ24}. Generally, the weights of neurons with high shortcut scores are set to 0 to achieve pruning~\cite{ali2024mitigating}. By quantifying the contribution of feedforward neural networks and attention heads to prediction results, some studies identify biased components of LLMs and alleviates their prompt brittleness by removing the corresponding components~\cite{Hanzhang2024}. However, the negative impact of component removal on LLMs is still debatable, as neurons have been observed to be polysemous, meaning that a single neuron can be activated by multiple terms~\cite{LiuKNMTW22, bills2023language}. So when shortcut components are removed, some corresponding useful knowledge may also be discarded, introducing new noise.

    \item \textbf{Calibration} corrects biased outputs of LLMs by modifying theirs probability distribution~\cite{GuoPSW17}. Contextual Calibration obtains context-driven biased distributions by replacing the sample to be predicted with meaningless inputs, such as `N/A'~\cite{ZhaoWFK021}. In an ideal unbiased environment, even if the sample to be predicted is set to `N/A', LLMs should yield an output similar to 50\% Positive and 50\% Negative. If the prediction distribution shifts, then it can be attributed to the corresponding context. Subsequently, during actual predictions, LLMs' outputs are adjusted based on the biased distribution to mitigate shortcut learning. 

    As research progresses gradually, Contextual Calibration has spawned various calibration methods tailored for different shortcut mitigation scenarios. Prototypical Calibration employs a Gaussian mixture distribution to estimate prototypical clusters for all categories and then utilizes the likelihood of these prototype clusters to calibrate LLMs' predictions~\cite{han2022prototypical}. Domain-context Calibration employs randomly selected in-domain words from the task corpus to estimate the label bias of the language model and utilizes this estimation for bias mitigation~\cite{FeiHCB23}. Similarly, PMI$_{DC}$ re-weighs scores based on the likelihood of hypotheses (answers) given premises (questions) in a specific task domain~\cite{HoltzmanWSCZ21}.

    But these calibration schemes may fail due to their inability to effectively estimate contextual bias with only content-free and in-domain random tokens. To address this, Batch Calibration updates the bias as more batches of input data are processed, allowing the bias to stabilize after multiple mini-batches are considered~\cite{ZhouWPMCHR24}. Generative Calibration adjusts the margin of the labels estimated via Monte-Carlo sampling over the in-context model to simply calibrate the contextual prediction distribution~\cite{JiangZLZL23}. In-Context Calibration calibrates the test distribution of the demonstration using the prior semantics expected from the samples~\cite{JangJKJY24}. PriDe is applied to the task of multiple choice questions by arranging the option content on a small number of test samples to estimate prior information, and then using the estimated prior information to debias the remaining samples~\cite{Zheng0M0H24}. Unlike the aforementioned probability-based calibration methods, NOISYICL calibrates language models by adding Gaussian noise to the parameters of LLMs, thereby reducing the prediction bias and unreliable confidence in ICL~\cite{Yufeng24}.
\end{itemize}

\subsection{Prompt-centric Approach}  \label{sec:Prompt-centric}
The prompt-centric approach aims to reduce LLMs' focus on shortcuts by modifying the prompts. Depending on the specific part of the prompt that is modified, it can be classified into three methods: shortcut-based method, instruction format-based method, and prompt search-based method.

\begin{itemize}[leftmargin=*,align=left]
    \item \textbf{Shortcut-based method}. It mitigates the unhealthy focus of LLMs on shortcuts by masking or perturbing shortcut words or concepts. Zhou et al. mask words with high relevance to shortcut concepts, where relevance is measured using Pointwise Mutual Information~\cite{ZhouX0A0H24}. Wang et al. replace entities that may rely on shortcuts with placeholders and alleviates the focus on a particular entity by guiding LLMs to understand the placeholders as a set of similar entities~\cite{WangMWZC23}. 

    \item \textbf{Instruction format-based method}. It guides LLMs to produce unbiased responses by modifying the format of the prompt or adding more detailed instructions. For example, it is advisable to utilize semantic verbalizers as much as possible to ensure that the label tokens have semantic relevance to the task~\cite{SiFJFC023}. Alternatively, in multiple-choice scenarios, the multiple options can be rearranged, and the rearranged results can be integrated using a majority voting method from LLMs to mitigate the risks associated with shortcuts related to option order~\cite{PezeshkpourH24, WangLCCZLCKLLS24}. Sun et al. recommend avoiding overloading the LLMs with too much task information that could lead to remembering previous false associations by mixing prompts and labels. Mixing prompts is achieved by substituting the first sample in the demonstration for mixed prompts, while mixed labels involves randomly replacing the labels with meaningless symbolic tags~\cite{SunXLJCZ24}. 

    In addition to the aforementioned modifications to prompt formats, some methods also use hints~\cite{QiLHWLWL23}, Chain-of-Thought (CoT)~\cite{yuan2024llms}, or causal graph~\cite{yamin2024failure} to guide LLMs in making more reasonable inferences. For example, Zhang et al. utilize the chain of thought generated by LLMs as a mediator variable, and adjust the causal relationship between the input prompt and the output answer through front-door adjustment to mitigate ICL bias~\cite{Congzhi2024}. Task-specific causal graphs can more reliably elicit the causal reasoning ability of language models and alleviate shortcuts caused by causal order conflicts~\cite{yamin2024failure}.

    \item \textbf{Prompt search-based method} aims to mitigate biases acquired by LLMs from context by searching for unbiased prompts. Unlike instruction format-based method, this approach focuses on retrieving reasonable prompts within a certain range, rather than having a clear modification target. Prompt search-based method can easily be associated with some recent retrieval-augmented approaches, which enhance the quality of ICL by retrieving samples from the corpus that are helpful for predictions~\cite{YoranWRB24, RamLDMSLS23}. Chen et al. observe a strong negative correlation between the sensitivity and accuracy of ICL, indicating that predictions sensitive to perturbations are unlikely to be correct~\cite{ChenZYM023}. Based on this, SENSEL is proposed to avoid providing users with incorrect predictions by discarding samples that are sensitive to perturbations. Lu et al. utilize the predicted label distribution statistics and propose an entropy-based metric to measure the quality of candidate prompts with diffenret order~\cite{LuBM0S22}. Gonen et al. assume that lower perplexity correlates with better performance and selects prompts with low perplexity as input~\cite{Gonen0BSZ23}.
\end{itemize}

\section{Rethinking of Shortcut learning}  \label{sec:thinking}
This section compares the shortcut learning in ICL with that in LPLMs, and outlines several potential future research directions.

\subsection{Differences from Shortcuts in LPLMs} \label{sec:diff}
Given that previous reviews have primarily focused on research into LPLMs, we deem it necessary to elucidate the similarities and differences between shortcut learning in LLMs and LPLMs to  bridge the two different studies. 

\subsubsection{Shortcut Types} \label{sec:diff1}

The difference between them primarily arises from the distinctions in research paradigms between LPLMs and LLMs. LPLMs typically require fine-tuning for different tasks, whereas LLMs unify various NLP tasks and only need a few samples to guide their inference. Therefore, ICL encounters a broader range of shortcut types from the few samples compared to those in LPLMs, even though both include shortcut types such as lexicon ~\cite{NivenK19}, overlap~\cite{LaiZFHZ21}, position~\cite{KoLKKK20}, and text style~\cite{QiCZLLS21}. For instance, since ICL do not require fine-tuning, the biases inherent in LLMs themselves can strongly influence various tasks, leading to the emergence of the instinctive shortcuts. Furthermore, since LLMs can easily infer answers from few samples, they naturally encounter shortcuts arising from these samples' group dynamics. The shortcuts related to group dynamics may, to some extent, resemble the challenges faced in long-tailed classification, where models tend to learn answers from head classes rather than tail classes~\cite{LiMSSDWX23}. But besides the skew in categories, LLMs are also influenced by more complex group decision-making process, such as the parity of options~\cite{balepur2024artifacts}. This can be attributed to LLMs' stronger comprehension abilities, enabling them to uncover complex correlations that LPLMs are unable to detect with few samples.

\subsubsection{Mitigation Strategies} \label{sec:diff2}

Another notable difference lies in shortcut mitigation strategies. For LPLMs, a widely adopted mitigation approach involves adjusting the model during fine-tuning, which encompasses techniques such as adversarial training~\cite{RashidLR20}, regularization~\cite{StaceyBR22}, Product-of-Experts~\cite{Sanh0BR21}, reweighting~\cite{LiuHCRKSLF21}, contrastive learning~\cite{Rui23}, and other related operations. These fine-tuning based methods reformulate the parameters of the model to mitigate the model's focus on shortcuts or samples containing shortcuts. But for LLMs, fine-tuning to mitigate shortcuts is rare, and the primary methods focus on research based on forward propagation. But we can still find extensions of LPLMs in the related research on LLMs. For instance, Calibration methods are similar to Product-of-Experts to some extent, both of which adjust the prediction probabilities. Alternatively, shortcut-based methods are similar to keyword regularization methods in that they both apply specific tokens to a single sample~\cite{SONG2025103964}. This intrinsic connection may guide researchers in the direction of transferring mitigation methods from LPLMs to ICLs. 

\subsubsection{Shortcut Search}  \label{sec:diff3}

Based on the apriorism premise that the shortcut is known, it is easy to make actions to alleviate the shortcut for LPLMs. However, this requires expert knowledge, and its generalization to complex and variable applications is limited~\cite{HanT21}. Therefore, some methods explore scenarios where shortcuts are unknown and search for biased samples and shortcuts using knowledge distillation~\cite{UtamaMG20} and interpretable methods, such as attention scores~\cite{MoonMLLS21}, shortcut degree~\cite{DuMJDDGSH21}, saliency technique~\cite{SocherPWCMNP13}, integrated gradients~\cite{SundararajanTY17, ChoiJHH22}, counterfactual attribution~\cite{WangC20}, etc. These methods may also be used in the study of shortcut learning in ICL~\cite{JuCY0DZL24}. For example, Satyapriya et al. determine the facilitating effect of post hoc explanations on LLMs~\cite{KrishnaMSG0L23}. Besides, interpretable methods also help to search for higher quality samples, which again helps to alleviate shortcuts~\cite{Tai2024, PengDY00OT24}. Still, interpretable methods in shortcut learning still need to be further explored, as described in Section~\ref{sec:future}. 

\subsection{Future Directions} \label{sec:future}

\subsubsection{More Robust Evaluation Benchmark}  \label{sec:future1}
Studies have demonstrated that the correlation between the performance exhibited by the model and the prompts is significant, which suggests that the evaluation benchmark itself also exhibits bias\cite{SiskaMAB24}. When the model responds to certain similar prompts within the benchmark, it tends to produce similar results. Lorenzo et al. also indicate that LLMs can exploit shortcuts in the benchmark to obtain inflated evaluation scores~\cite{pacchiardi2024leaving}. In addition, data contamination and leakage can also impact the evaluation of LLMs' shortcuts, as LLMs tend to provide good prediction results directly for benchmarks that have appeared in the training corpus~\cite{Kun2023}. Nishant et al. also imply that leakage of the test set might have an impact on multiple-choice questions without the question~\cite{balepur2024artifacts}. The aforementioned studies suggest that caution should be exercised when interpreting the performance results of LLMs, and the impact of shortcuts should be fully considered. Furthermore, efforts should be made to minimize the potential influence of shortcuts on prediction results before constructing benchmarks. 

\subsubsection{More Shortcut Related Tasks}  \label{sec:future2}
Despite the extensive exploration of shortcut learning in ICL, given the diversity of NLP tasks, it remains crucial to expand the boundaries of research coverage and tasks. As shown in Table~\ref{tab:tasks}, the performance of some shortcut types on some tasks has not been fully explored and may contain new discoveries. Besides, on more tasks such as Table QA, only a small portion of the entire table is relevant to deriving the answer to a given question, and irrelevant information may become potential shortcuts that affect the prediction outcomes of LLMs~\cite{PatnaikCAB0K24}. For another example, in the planing task, which determines the process of the sequence of actions needed to achieve the goal, LLMs have also been observed to focus on short-horizon and low-level planning, which may also become a shortcut learning~\cite{xie2024revealing}. These expanded tasks are highly dependent on relevant benchmarks, so designing and constructing corresponding datasets and evaluation methods for new tasks also require attention. 

\subsubsection{More Interpretability}  \label{sec:future3}
Although the studies in Section~\ref{sec:cause} explore the causes of shortcuts from different aspects, most of them are empirical summaries based on experimental observations, lacking an interpretation of the causes of shortcuts from an explainable perspective. For instance, how ICL influences the role of instinctive shortcuts in different NLP tasks, and at which step of the forward propagation in LLMs do these shortcuts exert their impact, is it during aggregation or distribution~\cite{WangLDCZMZS23}? Interpretable methods can enhance the intuitive presentation of shortcut learning in ICL and facilitate effective mitigation of shortcuts by addressing their underlying causes. 

However, there are still many limitations in applying these interpretable methods on LLMs. For example, the commonly used integrate gradient methods requires LLMs to perturb and assign values to each feature in the input. The time cost is unacceptable on large-scale datasets and LLMs. Similarly, the saliency technique requires performing at least one backpropagation on the LLMs to compute the saliency score for each element, which is not applicable for black-box LLMs~\cite{WangLDCZMZS23}. Therefore, the time cost and the opacity of black-box LLMs still need to be considered in the interpretable study of shortcut learning. A feasible approach is to attribute LLMs through small-scale models~\cite{KrishnaMSG0L23}, but this ignores the differences between small-scale models and LLMs~\cite{Luo2024RAG}. Distilling knowledge from LLMs into the small-scale models could be a potential solution~\cite{HaSLUY21}. Another viable way is to reduce the complexity of interpretable evaluation through heuristic search methods, but this may also lead to suboptimal results~\cite{LiQ23a}. For black-box LLMs, the self-explanatory ability of LLMs can be a shortcut learning exploration tool in this process~\cite{Shiyuan2023}. All in all, interpretable analysis in shortcut learning still needs to be further explored in ICL.

\subsubsection{More Discussion in Shortcut Unknown Scenarios}  \label{sec:future4}
In the research of LPLMs, shortcut identification is a crucial preliminary step for shortcut mitigation~\cite{DuHZTH24}, as in many cases, the types of shortcuts in tasks are unknown, or multiple shortcuts coexist. But in ICL, most studies assume that LLMs are vulnerable to shortcuts, or simply presume that the types of shortcuts are known and explore methods to assess performance under these known shortcuts. This idealized and strong assumption of known shortcuts may fail when facing new tasks, leading to limitations in shortcut mitigation solutions. Therefore, exploring how LLMs identify and utilize shortcuts in tasks where the shortcuts are unknown is also one of the subsequent research directions. In this scenario, interpretable methods may be used as a pre-technical means for shortcut mining.

\subsubsection{Decoupling of Instinctive and Acquired Shortcuts}  \label{sec:future5}
Based on the discussion in Section~\ref{sec:cause}, instinctive shortcuts are established during the training process of LLMs, while acquired shortcuts originate from demonstrations in the inference process. Instinctive shortcuts should emerge earlier in the lifecycle of LLMs. Consequently, research solely focused on acquired shortcuts may be potentially influenced by instinctive shortcuts, leading to biased conclusions. Jannik et al. show that label relationships inferred from pretraining appear to have permanent effects that can not be overcome by observations in context~\cite{KossenGR24}. When the LLMs are unfamiliar with the task, it is difficult to solve the problem by recalling the pre-trained knowledge. In this case, only similar examples can improve model performance by providing more direct shortcuts~\cite{HuTYZ24}. Moreover, in other application scenarios, such as Retrieval Augmented Generation, LLMs has a strong bias towards leveraging contextual information to answer questions, while minimally relying on internal knowledge~\cite{Hitesh2024}, again suggesting that LLMS may be more affected by acquired shortcuts. Therefore, there is a need to strengthen the exploration of how these two different types of shortcuts interact with each other. Furthermore, in the research on analyzing and mitigating acquired shortcuts, it is important to consider reducing or eliminating the influence of instinctive shortcuts to ensure the objectivity of the experimental results.

\subsubsection{Exploration of Multiple Shortcut Coexistence}  \label{sec:future6}
This can be regarded as an extension of the decoupling of instinctive and acquired shortcuts. Li et al. show that mitigating one shortcut increases the dependence on other shortcuts~\cite{LiEGHHCXI23}. So, when multiple shortcuts coexist, quantifying the mutual influence between shortcuts and giving the corresponding mitigation plan is also one of the future research directions. Some technical gains of Intersectional Debias can guide the corresponding research methods, such as nullspace projection~\cite{SubramanianHBCF21}. But in complex scenarios, when the given shortcut is unknown, the corresponding research faces more challenges. 

\section{Conclusion}
This paper comprehensively reviews the existing literature related to shortcut learning in LLMs, conducts a thorough classification of shortcuts in ICL, discusses their causes, presents existing evaluation benchmarks and mitigation strategies, and identifies key challenges and potential directions for future work. To our knowledge, this is the first review dedicated to the phenomenon of shortcut learning in ICL. We aim to emphasize the current research status of shortcut learning in the context of LLMs and provide insights to guide future work.

\Acknowledgements{This work was supported by China Postdoctoral Science Foundation Funded Project (No. 2024M761122), the National Natural Science Foundation of China (No.62476111), and the Department of Science and Technology of Jilin Province, China (No.20230201086GX)}



\clearpage
\bibliographystyle{splncs04_my}
\bibliography{custom}
\end{document}